\documentclass[lettersize,journal]{IEEEtran}
\usepackage{amsmath,amsfonts}
\usepackage{algorithmic}
\usepackage{algorithm}
\usepackage{amssymb}
\usepackage{amsmath}   %0527
\usepackage{array}
\usepackage[caption=false,font=normalsize,labelfont=sf,textfont=sf]{subfig}
\usepackage{textcomp}
\usepackage{stfloats}
\usepackage{url}
\usepackage{verbatim}
\usepackage{graphicx}
\usepackage{cite}
\usepackage[pagebackref,breaklinks,colorlinks]{hyperref} % 点击参考文献跳转
\begin{document}

\bstctlcite{IEEEexample:BSTcontrol}

\title{Model-Contrastive Federated Domain Adaptation}

\author{Chang'an Yi (yi.changan@fosu.edu.cn), Haotian Chen, Yonghui Xu, Yifan Zhang
        % <-this % stops a space
% \thanks{This paper was produced by the IEEE Publication Technology Group. They are in Piscataway, NJ.}% <-this % stops a space
% \thanks{Manuscript received April 19, 2021; revised August 16, 2021.}}
}

% The paper headers
\markboth{Journal of \LaTeX\ Class Files,~Vol.~14, No.~8, May~2023}%
% \markboth{Please submit the manuscript to the Special Issue on Trustworthy Federated
% Learning}%
{Shell \MakeLowercase{\textit{et al.}}: A Sample Article Using IEEEtran.cls for IEEE Journals}

% \IEEEpubid{0000--0000/00\$00.00~\copyright~2021 IEEE}
% Remember, if you use this you must call \IEEEpubidadjcol in the second
% column for its text to clear the IEEEpubid mark.

\maketitle

\begin{abstract}
Federated domain adaptation (FDA) aims to collaboratively transfer knowledge from source clients (domains) to the related but different target client, without communicating the local data of any client. Moreover, the source clients have different data distributions, leading to extremely challenging in knowledge transfer. Despite the recent progress in FDA, we empirically find that existing methods can not leverage models of heterogeneous domains and thus they fail to achieve excellent performance. In this paper, we propose a model-based method named FDAC, aiming to address {\bf F}ederated {\bf D}omain {\bf A}daptation based on {\bf C}ontrastive learning and Vision Transformer (ViT). In particular, contrastive learning can leverage the unlabeled data to train excellent models and the ViT architecture performs better than convolutional neural networks (CNNs) in extracting adaptable features. To the best of our knowledge, FDAC is the first attempt to learn transferable representations by manipulating the latent architecture of ViT under the federated setting. Furthermore, FDAC can increase the target data diversity by compensating from each source model with insufficient knowledge of samples and features, based on domain augmentation and semantic matching. Extensive experiments on several real datasets demonstrate that FDAC outperforms all the comparative methods in most conditions. Moreover, FDCA can also improve communication efficiency which is another key factor in the federated setting.
\end{abstract}


\section{Introduction}
\IEEEPARstart{F}{ederated} learning (FL)~\cite{kairouz2021advances, yang2019federated} enables different clients (e.g., companies and mobile devices) to jointly train a machine learning model since the data is usually dispersed among different clients in practice. Furthermore, no client is allowed to share its local data with any other client or the centralized server. However, a model trained with FL often fails to generalize to new clients (domains) due to the problem of domain shift \cite{yang2019federated}. For example, one client may contain pictures of mostly simulation environments, while another is mostly real environments. The phenomenon of domain shift has been thoroughly summarized in the survey of transfer learning \cite{pan2009survey, zhuang2020comprehensive}. In practice, federated domain adaptation (FDA) has become the main branch of FL, aiming to transfer knowledge from the decentralized clients to a different but related client (multi-source-single-target) \cite{peng2019federated, feng2021kd3a}, or from one client to decentralized clients (single-source-multi-target) \cite{yao2022federated}. FDA has gained wide attention in fields ranging from healthcare \cite{chen2020fedhealth, zhang2022two}, recommendation systems \cite{liu2021fedct}, Internet of Things \cite{mcmahan2017communication} to robotics \cite{yu2022towards}, due to the increasing data protection regulations and privacy concerns.

\begin{figure}[!t]
	\centering
	\includegraphics[width=6cm]{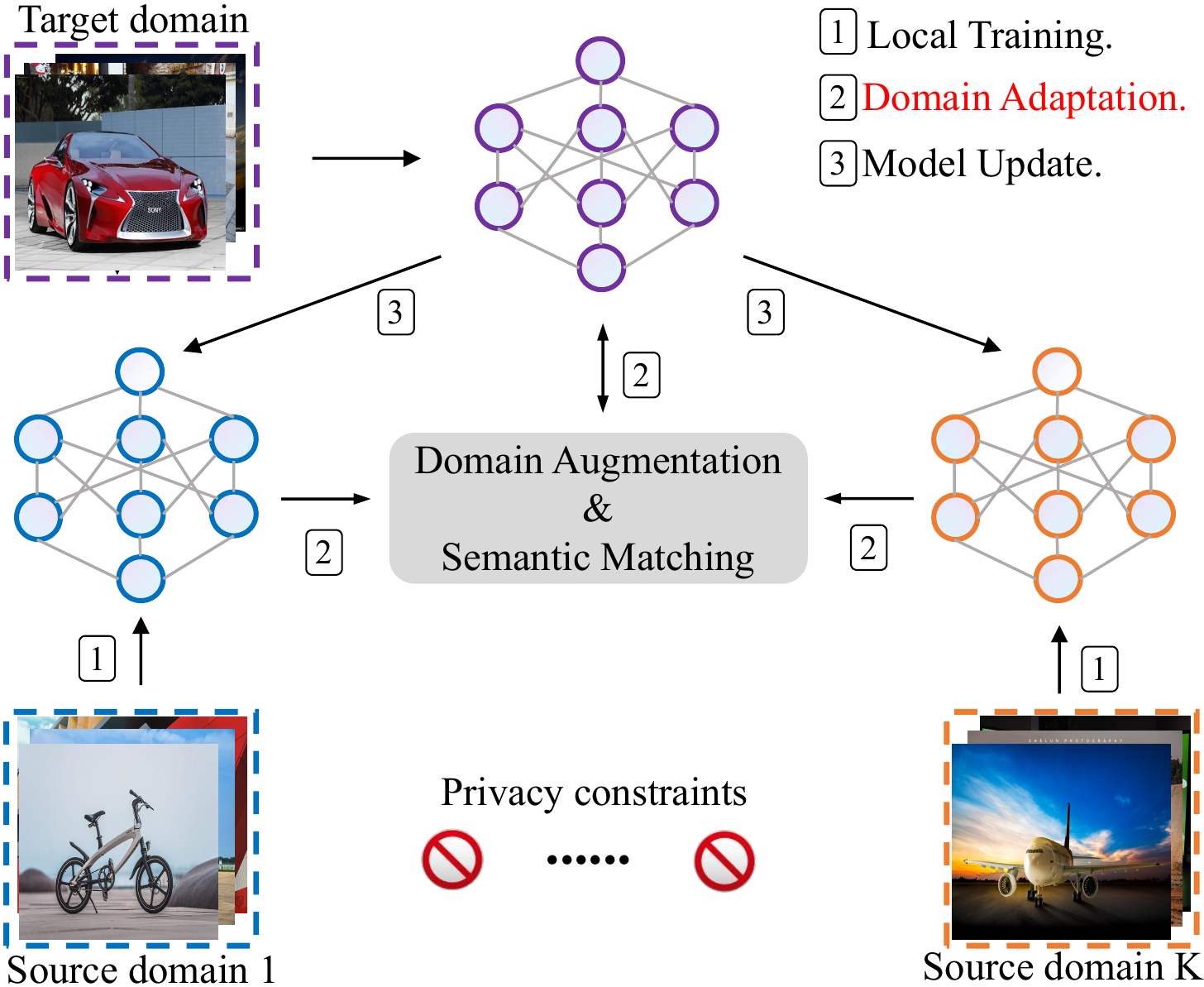}
	\caption{Illustration of our proposed FDAC method for federated domain adaptation. Domain augmentation and semantic matching are the key components to contrastively leverage different models at domain-level and category-level, respectively. Both the performance and communication efficiency are considered in this method. }
	\label{fig:illustration}
	\vspace{-0.5cm}
\end{figure}

The federated setting has some additional challenges \cite{peng2019federated}, in particular, the $\mathcal{H}$-divergence \cite{ben2010theory} can not be minimized due to privacy constraints. As a result, existing domain adaptation techniques \cite{tzeng2017adversarial, zhu2017unpaired, liu2019transferable, zhou2021domain} can not be applied in FDA. Some works \cite{peng2019federated, liang2020we, feng2021kd3a} attempt to adapt knowledge without accessing the source data, however, they fail to achieve high performance. Due to the heterogeneity of local data distribution across source domains, how to leverage the data-privacy source models and unlabeled target data becomes a main challenge. At least two problems should be considered in order to handle this challenge in FDA. Firstly, how to extract transferable features to adapt knowledge across heterogeneous domains? Secondly, how to align the conditional distributions by learning from the source models without accessing their local data?

\IEEEpubidadjcol  % 版权命令，放在第一页左边，不然会跟文字重叠
Vision Transformer (ViT) can extract more adaptable and robust features compared to traditional deep neural networks (DNNs) such as convolutional neural networks (CNNs) \cite{dosovitskiy2020image, sun2022safe, zhou2022understanding}. However, ViT-based methods face several challenges \cite{han2022survey}. For example, they heavily rely on large-scale training data. Thus, it is more difficult to bridge the large domain gap in a federated setting, due to the diversity of heterogeneous data. To solve this problem, domain augmentation is necessary to consider the complementarity among domains \cite{shu2021open}. According to \cite{verma2019manifold}, manipulating the hidden layers of DNNs can obtain better feature representations. Consequently, utilizing the latent architecture of ViT may augment data at domain-level and generate transferable features to bridge the domain discrepancy in FDA.

In recent years, contrastive learning has become a popular discriminative method based on embedding the augmented data and it has shown promising results on downstream tasks such as classification \cite{chen2020simple, chen2020simple, wang2022cross}. Since a prototype can represent a group of semantically similar samples \cite{snell2017prototypical}, the prototypes of each source domain can be generated based on the source models without accessing the local data \cite{liang2020we}. Then, the conditional distributions can be aligned by matching the semantic information of the source and target domains based on contrastive learning. Although several approaches \cite{singh2021clda, chen2021transferrable, tanwisuth2021prototype, wei2022multi, wang2022cross} have been proposed to learn transferable representations across domains based on contrastive learning or prototypes, these settings are relatively simpler than the setting under ViT and FL, since ViT is more data data-hungry than CNNs and the communication efficiency should be considered in the federated setting.

In this paper, we propose a model-aware contrastive approach (FDAC) to address {\bf F}ederated {\bf D}omain {\bf A}daptation based on {\bf C}ontrastive learning and Vision Transformer. In particular, FDAC considers the multi-source-single-target FDA setting \cite{peng2020federated, feng2021kd3a}, which is more popular than the single-source-multi-target scenario \cite{yao2022federated}. The general idea of FDAC is illustrated in Fig. \ref{fig:illustration}, where domain augmentation and semantic matching are two key components to adapt knowledge from different models. In summary, the main contributions of FDAC are presented as follows:   

\begin{enumerate}
     \item We utilize the hidden architecture of ViT to further explore the feature transferability among heterogeneous domains. To the best of our knowledge, this method is the first attempt to investigate transferable representations by manipulating the latent architecture of ViT under the federated setting. 
     
     \item We propose a novel framework integrating domain augmentation and semantic matching to adapt knowledge from all the source models. Moreover, this framework can increase data diversity, align class-conditional distributions across domains and avoid catastrophic forgetting.   

     \item We have performed extensive experiments on several real datasets to demonstrate the effectiveness of our proposed method FDAC. The comparative results indicate that FDAC consistently outperforms the state-of-the-art FDA approaches in most conditions. Moreover, FDAC can better improve communication efficiency which is also a key factor in FL.

     % \item We propose a novel framework integrating domain augmentation and semantic matching to make the target model learn from all the source models. Moreover, this framework can learn generalizable representations and align class-conditional distributions across domains. 
     
    % Compared to traditional contrastive learning methods where samples are augmented, FDAC is a higher-level method where blocks and a categories are augmented. 

    % Moreover, we also investigate the sensitivity of block selection during domain adaptation without violating the federated setting.
    % \item We use contrastive learning to reduce the distance of same category samples while pushing away different category samples. Moreover, we utilize an affinity matrix to minimize the distance of same categories and minimize the distances of different categories. In this process, all the source and target domains are considered simultaneously.  
\end{enumerate}

% \IEEEpubid{0000--0000/00\$00.00~\copyright~2021 IEEE}

The rest of this paper is organized as follows. Section II provides an overview of the related work. Section III describes the proposed FDAC framework in detail. Experimental results are reported and discussed in Section IV. Conclusions are presented in Section V.

\section{Related Work}
In this section, we review important research related to this work, including: (1) federated domain adaptation; (2) contrastive learning; and (3) Vision Transformer.  

\textbf{\begin{figure*}[!htbp]
	\centering
	\includegraphics[width=16cm]{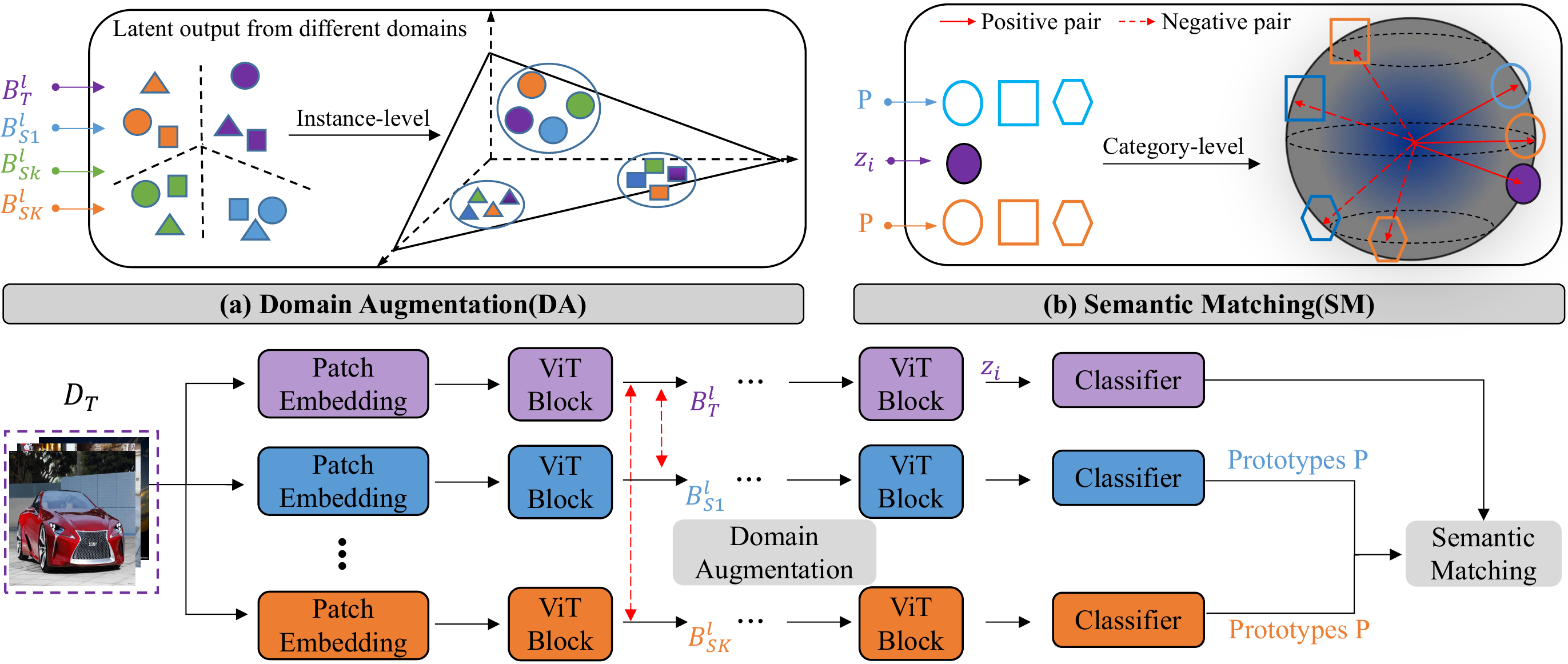}
	\caption{The general framework of our proposed method FDAC. This method adapts knowledge from heterogeneous models based on contrastive learning and ViT. Firstly, it manipulates the latent architecture of ViT and the augmented data originates from all the source models based on target samples. Secondly, it matches the semantic information across domains based on the prototypes of each source model and the pseudo labels of target samples.}
	\label{fig:framework}
	\vspace{-0.5cm}
\end{figure*}}

\subsection{Federated Domain Adaptation}
\cite{mcmahan2017communication} is the first work to propose the concept of federated learning (FL), which aims to bring collaborative machine learning opportunities for large-scale distributed clients with data privacy and performance guarantees. Then, many works attempt to extend the implementation mechanism \cite{gong2022preserving} or discuss it in real applications such as fairness \cite{}, robustness \cite{} and FDA \cite{peng2020federated}. The development of federated learning systems is believed to be an exciting research direction which needs the effort from system, data privacy and machine learning communities \cite{li2021survey}. In order to encourage the clients from actively and sustainably participating in the collaborative learning process, the research of ensuring fairness in FL is attracting a lot of interest \cite{}. Since FL systems are vulnerable to both model and data poisoning attacks, \cite{} provides a broad overview of existing attacks and defenses on FL. Based on the distribution characteristics of the data, federated learning can be classified into vertically federated learning, horizontally federated learning and federated transfer learning \cite{yang2019federated}. 

FADA \cite{peng2020federated} extends unsupervised adversarial knowledge transfer to the constraints of federated learning, however, the communication cost of FADA is huge which will cause privacy leakage. KD3A \cite{feng2021kd3a} is robust to communication rounds based on knowledge distillation and vote-based pseudo labels. Similar to KD3A \cite{feng2021kd3a}, pseudo labeling is also used in SHOT \cite{liang2020we} which only needs well-trained source models. Adversarial training is often used in centralized learning to mitigate bias, since the heterogeneous data may yield unfair and biased models. \cite{hong2021federated} considers adversarial training in the federated setting, and it can output a debiased and accurate model. Different from these FDA scenarios that can be categorized as multi-source-single-target, \cite{yao2022federated} handles the single-source-multi-target scenario. The key challenge of FDA is to adapt knowledge from heterogeneous models, while obeying regulations and policies to protect privacy.

\subsection{Contrastive Learning}
Contrastive learning has become the most popular style of self-supervised learning in fields such as computer vision and natural language processing, since it can avoid the cost of annotating large-scale datasets \cite{jaiswal2021survey}. Different from generative methods, contrastive learning is a discriminative approach that aims to embed the augmented versions of the positive samples close to each other while trying to push away embeddings from negative samples. In this way, generative and contrastive approaches can be integrated to utilize the unlabeled samples to learn the underlying representations \cite{wei2022multi, chen2022contrastive}.

\cite{zhang2021unleashing} applies a hard pair mining strategy to enhance contrastive fine-tuning since the hard pairs are more informative and challenging. Several works attempt to apply existing self-supervision techniques to ViT. \cite{chen2021empirical} investigates the effects of training self-supervised ViT and finds that instability is a major issue. \cite{caron2021emerging} finds that self-supervised pretraining in a standard ViT model achieves similar or better performance compared to the best CNNs specifically designed for the same setting. Different from the above methods, \cite{yun2022patch} can utilize the architectural advantages of ViT and learn patch-level representation. Since the instance invariance assumption can be easily generalized to domain adaptation tasks, \cite{chen2021transferrable} finds that contrastive learning is intrinsically a suitable candidate for domain adaptation, where both transferability and discriminability are guaranteed. However, as far as we are concerned, very few works attempt to address FDA by simultaneously considering all the domains based on contrastive learning.

\subsection{Vision Transformer}
Self-attention mechanism is base component in Vision Transformer (ViT). ViT has fewer parameters and the training process converges more quickly compared to CNNs. Both of these advantages are important in the federated setting \cite{kairouz2021advances}. A ViT  model directly applies a pure Transformer to image patches to classify full samples \cite{vaswani2017attention, dosovitskiy2020image}. The self-attention mechanism of ViT connects every patch token with the classification and the potential of ViT has inspired many new approaches. \cite{he2022transfg} is a fine-grained visual classification framework to investigate the potential of ViT, where the discriminative ability of classification tokens is also guaranteed based on contrastive loss. \cite{ma2022visualizing} introduces a quantification indicator to visualize and interpret the patch-level interactions in ViT. Different from pure ViT-based approaches, \cite{lin2022cat} proposes a cross-attention mechanism to integrate CNNs and Transformers to build a robust backbone, indicating that ViT and CNNs can complement each other through global connection and local connection. Since ViT has exhibited strong capability in learning robust representations, \cite{zhou2022understanding} systematically examines the role of self-attention and verifies it as a contributor to the improved robustness of ViT. ViT also works well in the field of segmentation. For example, \cite{} investigated the feasibility of using transformer-based deep architectures for medical image segmentation tasks and it also introduces an extra control mechanism in the self-attention module to extend the existing architectures. 

Although ViT has been successfully applied in tasks such as video processing and computer vision, the configurable architecture of ViT has not yet been fully explored, which might bring fine-grained model adaptation, especially in FDA where the source data can not be accessed directly.

The works most related to our proposed FDAC framework are transferable contrastive learning approaches proposed in \cite{chen2021transferrable, wei2022multi}. However, these works differ from FDAC in two aspects. Firstly, there backbones are both CNNs while the backbone of FDAC is ViT. Furthermore, FDAC manipulates the latent architecture of its backbone to align the data distributions in a fine-grained manner. Secondly, different from \cite{chen2021transferrable}, the augmented data of FDAC is the original data of each source domain. Different from \cite{wei2022multi}, each local source domain model of FDAC is trained only on its own data.

% The key difference between the above methods and our proposed FDAC is mainly that our work is the first attempt to utilize the latent architecure of ViT to address FDA, while the communication efficiency is also guaranteed. Furthermore, we will handle the new problems emerged when the backbone is ViT instead of traditional CNNs.

\section{The Proposed FDAC Framework}
% This section presents the proposed method to address the domain discrepancy between the multi-source and target domains without data sharing.
\subsection{Notations and Problem Statement}
We use $\{\mathcal{D}_{S_k}\}_{k=1}^K$ and $\mathcal{D}_T$ to denote the $K$ decentralized source domains and target domain, respectively. $\mathcal{D}_{S_k}$ contains $N_k$ labeled samples, i.e., $\mathcal{D}_{S_k} = \{(x_i^k, y^k_i)\}_{i=1}^{N_k} (1\leqslant{k}\leqslant{K})$. $\mathcal{D}_T$ has $N_T$ unlabeled samples, i.e., $\mathcal{D}_T = \{(x_i^T)\}_{i=1}^{N_T}$. Under our FDAC setting, the marginal data distributions of any source and target domains are different (i.e., $P_{S_k}{(x) \neq P_T{(x)}}$, $P_{S_k}{(x) \neq P_{S_j}{(x)}}$) while their conditional distributions are the same (i.e., $P_{S_k}{(y|x) = P_T{(y|x)}}$) $(1\leqslant{k, j}\leqslant{K})$. Each source domain can train a local model based on its own data and the model parameters can be communicated among domains. 

The goal of FDA is to learn a classifier for $\mathcal{D}_T$ under the privacy restrictions. To achieve that goal, there exists the following challenges:  
\begin{enumerate}
	\item It is challenging to increase the diversity of the target data without accessing the local data of each source domain.
	\item Since each category information can not be described in detail, it is challenging to align the conditional data distributions across different domains.
    % \item Since iteratively update models need a lot of communications among domains, it is challenging to guarantee the communication efficiency.
\end{enumerate}

\subsection{Overall Framework}
To achieve these challenges, we propose the FDAC method. The framework of FDAC is displayed in Fig \ref{fig:framework}, which aims to transfer knowledge from the different source models to the target model while the communication efficiency is also guaranteed. The implementation of FDAC is based on domain augmentation and semantic matching, corresponding to domain-level and category-level contrastive learning, respectively. Different from traditional transferable features learning \cite{chen2021transferrable, wei2022multi}, we utilize the configurable architecture of ViT to perform contrastive learning based on domain augmentation, since the latent manipulation of DNNs can improve feature representations \cite{verma2019manifold}. Moreover, this kind of domain augmentation can increase the data diversity of the target domain by complementing from each source domain. On the other hand, in order to exploit the class similarities to make knowledge transfer from source data to similar target categories, we extract domain-invariant features based on semantic matching. Since no source data is available to train the target model, we first generate prototypes for the source domains and then learn discriminative information based on those prototypes. Thus, these two components are also able to avoid catastrophic forgetting when knowledge is leveraged to adapt from different sources to the target domain. 

\subsection{Model-Contrastive Domain Augmentation}
The statistical learning theory \cite{vapnik1999nature} suggests that the model capacity and the diversity of the training data can characterize the generalization of a machine learning model. Inspired by \cite{shu2021open}, increasing the data diversity of multiply domains can enhance the generalization of representations. Due to the heterogeneity of local data in the federated setting, transferable feature representations are critical to enabling source models to make similar predictions based on semantically identical data. Motivated by this idea, we expand the diversity of target samples by augmenting data at domain-level. We observe that the target domain contains distinct knowledge  but lacks domain knowledge of other source domains. Our insight is to conduct domain augmentation on domain-level to increase the diversity of target data based on all the source domains. Moreover, the target domain is compensated with missing knowledge of classes and features from each source domain.

\textbf{The Backbone of ViT}. The backbone of ViT is, in essence, one kind of DNNs. Thus, the extracted features of the first blocks are relatively transferable, compared to the output features of the later blocks which are relatively discriminative. ViT can be used beyond a feature extractor since each block is independent and the output feature of any block can be fetched. Usually, an input sample of the ViT backbone is first divided into 196 patches with the fixed size $16*16$ \cite{zheng2022prompt}. The encoding layer converts the input patches into patch tokens, and then the positional embeddings are added to them. The input to the Transformer is the encoded patch tokens plus a classification token, denoted by $B^0$. The Transformer encoder consists of $L$ layers of Multi-head Self-Attention (MSA) and Multi-layer Perceptron (MLP)  blocks. Then, the output of the $l$-th $(1\leqslant{l}\leqslant{L})$ layer can be written as:
\begin{equation}
    \label{eq:vit1}
    \hat{B}^l = \text{MSA} \left(\text{LN}\left( B^{l-1}\right) \right) + B^{l-1},
    \begin{aligned}
    \end{aligned}
\end{equation}
\begin{equation}
    \label{eq:vit2}
    B^l = \text{MLP}\left(\text{LN}\left( \hat{B}^l\right) \right) + \hat{B}^l,
    \begin{aligned}
    \end{aligned}
\end{equation}
where $\text{LN}(\cdot)$ represents the layer normalization operator.

\textbf{Domain Augmentation}. Inspired by the configurable architecture of ViT, we design a transferable contrastive learning module in FDAC, based on domain-level data augmentation. The detail of this module is further illustrated in Fig. \ref{fig:framework}.a in detail. Given any target sample $x_i$, we can easily get the output of each domain, i.e., $\tilde{\mathrm{B}}^l_{i(k)}$ for the $k$-th  source domain and $\mathrm{B}^l_{i(T)}$ for the target domain of the $l$-th layer, respectively. Our goal is to minimize the data discrepancy between $\mathrm{B}^l_{i(T)}$ and $\tilde{\mathrm{B}}^l_{i(k)}$ from the same sample relative to that discrepancy from different samples. Assuming that the features are $\ell_2$-normalized, the domain-augmented contrastive loss is computed by:
\begin{equation}
    \label{eq:layer}
    \begin{aligned}
    % \mathcal{L}_{layer} = - \frac{1}{n} \ \sum_{i=1}^{N_T}{\sum_{\tilde{\mathrm{b}}^l_i \sim Sk}^{K}{\log \frac{e^{{\mathrm{b}^l_i}^{\mathrm{T}}}}}{}}
    \mathcal{L}_{DA}  = - \frac{1}{N_T}\sum_{i=1}^{N_T}\sum_{k=1}^K {\log \frac{ e^{ \left({\mathrm{B}^l_{i(T)}}^{\top} \tilde{\mathrm{B}}^l_{i(k)} / \tau \right) }}{\sum_{\mathrm{B}^l_k \sim A_i} e^{ \left( {\mathrm{B}^l_i}^{\top} \mathrm{B}^l_k/\tau \right) } },} 
    \end{aligned}
\end{equation}
where $\tau$ is a temperature hyper-parameter and $A_i$ denotes the negative pairs representing that the input target sample is not $x_i$. Since the feature extraction of sample $x_i$ is based on each source model, computing the output of the given block also follows the privacy-preserving policy of the federated setting.    

As indicated in Fig \ref{fig:framework}.a, our contrastive learning based on latent feature space is different from traditional contrastive learning in domain adaptation, since our contrastive mechanism can leverage knowledge from different sources and the augmented samples are originated from the source domains instead of the original target samples. According to Eq. (\ref{eq:layer}), the transferable representations are learned based on all the domains.  

\begin{algorithm}[!h]     
	\caption{FDAC Algorithm}
	\label{ag:algorithm}
	\begin{algorithmic}[1]
		\REQUIRE  Source domains $\{\mathcal{D}_{S_k}\}_{k=1}^K$ $(1\leqslant{k}\leqslant{K})$. Target domain $\mathcal{D}_T$.  
		\ENSURE  Target model $\mathcal{M}_T$.
        \WHILE{not converged}
		\STATE // Stage 1: Locally training for each source domain.
% 		\FOR{ $\mathcal{D}_{S_k}$  in $\mathcal{S}$ }
        
		\STATE Train $\mathcal{M}_{S_k}$ with classification loss by Eq. 
 (\ref{eq:sourcetraining}). 
% 		\ENDFOR
		\STATE // Stage 2: Adaptation on the target domain.
        \STATE \# Domain Augmentation:
        \STATE  Compute the loss $\mathcal{L}_{DA}$ according to  Eq. 
 (\ref{eq:layer}).
        \STATE \# Get prototypes from source domains:
        \STATE  $\mathbf{P} \leftarrow \text{Prototype Generation}$.
        \STATE \#  Semantic Matching:
        \STATE  Compute the loss $\mathcal{L}_{SM}$ according to Eq. 
 (\ref{eq:semantic}).
        
        % \STATE // Prototype Affinity:
        % \STATE  Compute the loss of Affinity Matrix based on Eq.(\ref{eq:affinity}).
        % \STATE $\mathbf{M}_a \gets $ PrototypeAffinity$\left(\{\mathbf{P}_{Sk}\}_{k=1}^{K}, \mathbf{P}_T \right)$.

        \STATE Train $\mathcal{M}_T$ with  Eq. (\ref{eq:obj}).
        % \STATE \# Model Aggregation:
        \STATE // Stage 3: Model Aggregation:
        \STATE $\mathcal{M}_T \gets \left(\sum_{k=1}^{K}\mathcal{M}_{S_k}, \mathcal{M}_T \right) $
        \STATE Return $\mathcal{M}_T$.
        \ENDWHILE
	\end{algorithmic}
\end{algorithm}

\subsection{Model-Contrastive Semantic Matching}
In federated learning, the source data is kept locally and the target data is unlabeled. Thus, it is extremely necessary for the target model to learn from all the locally-trained source models. In FDAC, we propose the category-level contrastive learning module as illustrated in Fig. \ref{fig:framework}.b. This module can align the data distributions through two steps. Firstly, it generates prototypes for each category of all the source domains. Secondly, it utilizes contrastive learning to minimize the distances of target samples to the source prototypes with the same classes relative to those with different categories. Moreover, pseudo labels are used in the second step since the target samples are unlabeled.

\textbf{Prototype Generation}. By exploring the supervised semantic information of multiple heterogeneous domains, we seek to generate domain-invariant prototypes for each category in each source domain. Inspired by \cite{saito2019semi}, the direction of a prototype should be representative of the features belonging to the corresponding category. Assume that each model $\mathcal{M}$ consists of a feature extractor $\mathcal{F}$ which is actually the backbone of ViT, and a classifier $\mathcal{C}$. We perform $\ell_2$-normalization on $\mathcal{F}$ and then use it as the input of $\mathcal{C}$ which consists of weight vectors $\mathbf{P} = [\mathbf{p}_1, \mathbf{p}_2, \cdots,  \mathbf{p}_C]$, where $C$ represents the number of categories. $\mathcal{C}$ takes $\frac{\mathcal{F}(x)}{||\mathcal{F}(x)||_2}$ as input and it outputs the probability $\mathcal{C}(x) = \sigma \left( \frac{\mathbf{P} \mathcal{F}(x)}{||\mathcal{F}(x)||_2} \right)$, where $\sigma$ is the softmax function. In sum, the prototype generation of source domain $k$ $(1\leqslant{k}\leqslant{K})$ is defined as:

\begin{equation}
    \label{eq:sourcetraining}
    \begin{aligned}
    \mathcal{L}_{S_k} \left(\mathcal{M}_{S_k}; \mathcal{D}_{S_k}\right) = -\underset{\left(x,y \right) \sim \mathcal{D}_{S_k}}{\mathbb{E}} \sum q \log \mathcal{M}(x), 
    \end{aligned}
\end{equation}
where $q$ is the one-hot encoding of the label. Then, we can use $\mathbf{P}$ to provide semantic guidance for the target model.

\textbf{Cross-domain Semantic Matching}. The true labels of the target domain are unavailable, thus, we first use pseudo labeling presented in \cite{feng2021kd3a} to produce high-quality pseudo labels $\tilde{y}_T$. We also use the generated pseudo labels to reduce the feature distribution gap by:

\begin{equation}
    \label{eq:pseudolabels}
    \begin{aligned}
    \mathcal{L}_{T} \left(\mathcal{M}_{T}; \mathcal{D}_{T}\right) = -\underset{\left(x,\widetilde{y}_T \right) \sim \mathcal{D}_{T}}{\mathbb{E}} \sum q \log \mathcal{M}(x),
    \end{aligned}
\end{equation}
where $\mathcal{M}_{T}$ represents the model of $\mathcal{D}_T$ and $q$ is the one-hot encoding of $\tilde{y}_T$. For a target sample $x$, we use an additional two-layer MLP $\mathcal{G}$ to obtain $\ell_2$-normalized contrastive features $z_i =\frac{\mathcal{G}(x)}{||\mathcal{G}(x)||_2} $, since a nonlinear projection can improve the performance of contrastive learning. Then, we use the supervised contrastive loss for adaptation. For a given target sample $x$, we take the prototypes with the same category as positive pairs $A_p$ and those with different classes as negative pairs $A_n$, according to the pseudo label of $x$. The cross-domain semantic matching loss $\mathcal{L}_{SM}$ is defined as: 
\begin{equation}
    \label{eq:semantic}
    \begin{aligned}
    \mathcal{L}_{SM} =  - \frac{1}{N_T}\sum_{i=1}^{N_T}\frac{1}{|A_p|}\sum_{\mathbf{p}_j \sim A_p } \log \frac{e^{\left( z_i^{\top} \mathbf{p}_j  \right)}}{\sum_{\mathbf{p}_k \sim A_n} e^{\left( z_i^{\top} \mathbf{p}_k  \right)} .} 
    \end{aligned}
\end{equation}

Both Eq. \eqref{eq:layer} and Eq. \eqref{eq:semantic} indicate that they can also avoid catastrophic forgetting when knowledge is contrastively transferred from multiply source models to the target model. In sum, the optimization problem for our FDAC approach is defined as:
\begin{equation}
    \label{eq:obj}
    \underset{\mathcal{M}_T}{\min} \quad \lambda_1 \mathcal{L}_{DA} + \lambda_2 \mathcal{L}_{SM} + \mathcal{L}_{T} , 
    \begin{aligned}
    \end{aligned}
\end{equation}
where $\lambda_1$ and $\lambda_2$ are hyper-parameters. We summarize the detailed training procedure of FDAC in Alg. \ref{ag:algorithm}, where the final model of the target domain is gained based on aggregation \cite{mcmahan2017communication}.

\subsection{Theoretical Analysis of FDAC}
This subsection performs theoretical analysis of the proposed FDAC method and demonstrates that its loss functions have regularization effectiveness and optimization effectiveness based on the theory of contrastive learning\cite{zhang2021unleashing, boudiaf2020unifying}.

For the domain augmentation loss $\mathcal{L}_{DA}$ proposed in Eq. (\ref{eq:layer}), we can get: 
\begin{equation}
    \label{eq:theorem1}
    \begin{aligned}
    \mathcal{L}_{DA} \propto \mathcal{H}\left( Z| \mathcal{M_T} \left( x \right)\right) - \mathcal{H} \left( Z \right), 
    \end{aligned}
\end{equation}
where $Z$ is the embedding features from both source and target domains. Eq. (\ref{eq:layer}) shows that $\mathcal{L}_{DA}$ significantly improves feature representations. Minimizing $\mathcal{L}_{DA}$ is equivalent to simultaneously minimize $\mathcal{H}\left( Z| \mathcal{M_T} \left( x \right)\right)$ and maximize $\mathcal{H} \left( Z \right)$. Minimizing $\mathcal{H}\left( Z| \mathcal{M_T} \left( x \right)\right)$ encourages the model $\mathcal{M_T}$ to generate low entropy clusters in the feature space for each given $x$ based on all domains. On the other side, maximizing $\mathcal{H} \left( Z \right)$ tends to learn a high-entropy feature space in order to increase the diversity for stronger generalization \cite{szegedy2016rethinking}.

For the semantic matching loss $\mathcal{L}_{SM}$ proposed in Eq. (\ref{eq:semantic}), we can get the infimum taken over classifiers: 
\begin{equation}
    \label{eq:theorem2}
    \begin{aligned}
        \mathcal{L}_{SM}  &\propto \mathcal{H}\left( Y| Z\right) - \mathcal{H} \left( Y \right) = - \mathcal{I}\left( Z; Y \right)\\ 
    &= \inf \mathcal{H}\left( Y; \mathcal{M}\left(x\right)|Z\right) - \mathcal{H} \left( Y\right), 
    \end{aligned}
\end{equation}
where $\mathcal{I}$ represents mutual information and $\mathcal{H} \left( Y\right)$ is a constant which can be ignored. Thus, minimizing $\mathcal{L}_{SM}$ with class prototypes will minimize the infimum of conditional cross-entropy $\mathcal{H}\left( Y; \mathcal{M}\left(x\right)|Z\right)$(i.e., mutual information maximization) provides an additional semantic guidance compared to pseudo labeling loss $\mathcal{L}_{T}$ with only cross-entropy. To sum up, FDAC can ......

\section{Experimental Evaluation}
In this section, we conduct extensive experiments to evaluate the performance of FDAC based on several publicly available datasets: DomainNet \cite{peng2019moment}, OfficeHome \cite{venkateswara2017deep}, OfficeCaltech \cite{fei2004learning}, PACS\cite{li2017deeper} and Cancer Dataset. Since the training of ViT heavily needs a lot of data, we do not select the dataset Digit-Five \cite{peng2020federated} which is relatively small. Furthermore, we also carry out experiments to demonstrate the advantage of our ViT-based augmentation compared to other ViT-based augmentation methods \cite{zhang2017mixup, sun2022safe}.\par 

\subsection{Datasets}
\begin{table}[!t]
\centering
\scriptsize
\caption{Experimental results on OfficeHome (Mean accuracy $\pm$ standard deviation) (\%)}
\label{tb:OfficeHome}
\begin{tabular}{c|cccc|c}
\hline
Method     & Art               & Clipart           & Product           & RealWorld         & Average           \\ \hline
ResNet50   & 65.9±0.1          & 49.7±0.1          & 76.5±0.8          & 79.1±0.3          & 67.8±0.3          \\
R50-Ours     & 73.0±0.5          & 60.2±0.3          & 83.6±0.2          & 84.4±0.7          & 75.3±0.4          \\
SourceOnly & 73.7±0.2          & 56.7±0.7          & 80.5±0.4          & 82.4±0.1          & 73.3±0.4          \\
PL         & 77.6±0.4          & 60.4±0.5          & 84.8±0.4          & 85.4±0.4          & 77.1±0.4          \\
SHOT       & 78.2±0.4          & 62.3±0.3          & 87.2±0.4          & 86.0±0.1          & 78.4±0.3          \\
FADA       & 76.7±0.6          & 60.5±0.5          & 81.2±0.6          & 83.4±0.4          & 75.4±0.5          \\
CPGA       & 75.2±0.2          & 61.5±0.4          & 82.8±0.3          & 83.1±0.4          & 75.7±0.3          \\
TransDA    & 76.4±0.3          & 56.5±0.4          & 84.4±0.6          & 86.4±0.4          & 75.9±0.2          \\
DECISION   & 77.8±0.3          & 62.9±0.4          & 87.3±0.1          & 85.0±0.2          & 78.3±0.3          \\
FADE       & 76.9±0.3          & 61.2±0.3          & 85.3±0.4          & 84.6±0.2          & 77.0±0.3          \\
KD3A       & 79.2±0.6          & 62.3±0.7          & 87.5±0.1          & 87.3±0.6          & 79.1±0.5          \\
Ours       & \textbf{80.2±0.1} & \textbf{65.3±0.5} & \textbf{89.2±0.1} & \textbf{88.6±0.1} & \textbf{80.8±0.2} \\ \hline
\end{tabular}
\end{table}

\begin{table}[!t]
\centering
\scriptsize
\caption{Experimental results on OfficeCaltech (Mean accuracy $\pm$ standard deviation) (\%)}
\label{tb:OfficeCaltech}
\begin{tabular}{c|cccc|c}
\hline
Method     & A                 & C                 & D                  & W                  & Average           \\ \hline
ResNet50   & 96.0±0.1          & 89.9±0.3          & 98.0±0.1           & 96.9±0.4           & 95.2±0.3          \\
R50-Ours     & 96.3±0.1          & 95.1±0.2          & 98.8±0.1           & 99.3±0.2           & 97.4±0.1          \\
SourceOnly & 96.0±0.2          & 94.1±0.3          & 98.7±0.0           & 97.3±0.7           & 96.5±0.3          \\
PL         & 96.0±0.5          & 94.7±0.5          & 99.4±0.5           & 98.8±0.5           & 97.2±0.5          \\
SHOT       & 96.0±0.1          & 96.4±0.2          & 99.5±0.5           & 99.4±0.3           & 97.8±0.3          \\
FADA       & 96.5±0.3          & 94.8±0.6          & 99.9±0.3           & 99.0±0.3           & 97.5±0.4          \\
CPGA       & 96.4±0.1          & 92.1±0.5          & 99.3±0.6           & 98.9±0.7           & 96.7±0.5          \\
TransDA    & 96.4±0.1          & \textbf{97.4±0.3}          & 100.0±0.0          & 98.8±0.2           & 98.2±0.2          \\
DECISION   & 96.3±0.1          & 96.9±0.3          & 99.5±0.1           & 99.8±0.2           & 98.1±0.5          \\
FADE       & 96.5±0.1          & 95.9±0.9          & 99.8±0.1           & 99.6±0.3           & 98.2±0.4          \\
KD3A       & 96.5±0.2          & 95.3±0.6          & 97.8±1.3           & 99.1±0.3           & 97.2±0.6          \\
Ours       & \textbf{96.9±0.1} & \textbf{97.0±0.4} & \textbf{100.0±0.0} & \textbf{100.0±0.0} & \textbf{98.5±0.1} \\ \hline
\end{tabular}
\end{table}
{\bf DomainNet.} DomainNet is the largest domain adaptation dataset containing about 0.6 million common images of 345 classes in 6 domains. The domains include clipart: collection of clipart images; real: photos and real world images; sketch: sketches of specific objects; infograph: infographic images with specific objects; painting artistic depictions of objects in the form of paintings and quickdraw: drawings of the worldwide players of game “Quick Draw!”.

\begin{table}[!t]
\centering
\scriptsize
\caption{Experimental results on PCAS (Mean accuracy $\pm$ standard deviation) (\%)}
\label{tb:PCAS}
\begin{tabular}{c|cccc|c}
\hline
Method     & P                 & C                 & A                 & S                 & Average           \\ \hline
Resnet50   & 97.4±0.1          & 63.2±1.0          & 82.2±0.9          & 68.6±0.3          & 77.9±0.6          \\
R50-Ours   & 98.5±0.2          & 91.1±0.2          & 93.9±0.4          & 86.1±0.6          & 92.4±0.4          \\
SourceOnly & 99.2±0.1          & 71.9±1.8          & 88.5±1.4          & 66.6±2.5          & 81.5±1.4          \\
PL         & 99.3±0.1          & 71.2±0.5          & 88.6±0.4          & 68.0±0.1          & 81.8±0.3          \\
SHOT       & 99.5±0.1          & 86.8±0.6          & 93.3±0.5          & 80.3±0.7          & 90.2±0.6          \\
FADA       & 99.0±0.1          & 82.8±0.5          & 91.2±0.5          & 81.2±0.7          & 90.2±0.5          \\
CPGA       & 98.8±0.2          & 81.8±0.4          & 89.5±0.3          & 75.9±0.6          & 86.5±0.4          \\
TransDA    & 99.6±0.2          & 84.4±0.9          & 94.0±0.1          & 78.1±0.2          & 89.0±0.4          \\
DECISION   & 99.6±0.2          & 86.4±0.3          & 94.9±0.6          & 73.6±0.1          & 88.6±0.4          \\
FADE       & 99.5±0.1          & 82.0±0.2          & 92.8±0.8          & 84.9±0.1          & 89.8±0.2          \\
KD3A       & 99.5±0.0          & 82.9±1.0          & 93.0±0.2          & 80.8±0.1          & 89.0±0.3          \\
Ours       & \textbf{99.8±0.0} & \textbf{90.1±0.3} & \textbf{95.9±0.1} & \textbf{87.8±0.4} & \textbf{93.4±0.2} \\ \hline
\end{tabular}
\end{table}

\begin{table*}[!t]
\centering
\scriptsize
\caption{Experimental results on breast cancer histology images classification of different modes. (Mean accuracy $\pm$ standard deviation) (\%)}
\label{tb:camelyon}
\begin{tabular}{c|ccccc|c}
\hline
Method     & A                 & B                 & C                 & D                 & E                 & Average           \\ \hline
Resnet50   & 88.7±0.4          & 85.1±0.6          & 81.0±0.2          & 87.4±0.7          & 75.0±0.1          & 83.4±0.4          \\
R50-Ours   &                   &                   &                   &                   &                   &                   \\
SourceOnly &                   &                   &                   &                   &                   &                   \\
PL         &                   &                   &                   &                   &                   &                   \\
SHOT       &                   &                   &                   &                   &                   &                   \\
FADA       &                   &                   &                   &                   &                   &                   \\
CPGA       &                   &                   &                   &                   &                   &                   \\
TransDA    &                   &                   &                   &                   &                   &                   \\
DECISION   &                   &                   &                   &                   &                   &                   \\
FADE       &                   &                   &                   &                   &                   &                   \\
KD3A       & 97.5±0.0          & 95.8±0.0          & \textbf{95.2±0.4} & 95.0±0.1          & 87.9±3.4          & 94.3±0.8          \\
Ours       & \textbf{98.1±0.0} & \textbf{96.4±0.1} & 92.9±0.5          & \textbf{95.3±0.9} & \textbf{94.6±1.0} & \textbf{95.5±0.5} \\ \hline
\end{tabular}
\end{table*}

\begin{table*}[!thp]
\renewcommand\arraystretch{1.3}
\scriptsize   
\centering     
\caption{Experimental results on DomainNet (Mean accuracy $\pm$ standard deviation) (\%)}
\label{tb:DomainNet}
\begin{tabular}{c|cccccc|c}
\hline
Method     & Clipart           & Infograph         & Painting          & Quickdraw                    & Real                         & Sketch                        & Average                      \\ \hline
Resnet101   & 58.7±0.1 & 20.7±0.1          & 47.8±0.2          & 10.0±0.1                     & 63.0±0.1                     & 46.7±0.2                      & 41.2±0.1                     \\
R101-Ours     & 71.0±0.1          & 22.3±0.4          & 57.5±0.3          & 17.2±0.3                     & 70.5±0.3                     & 57.5±0.1                      & 49.3±0.2                     \\
SourceOnly & 60.5±0.1          & 20.4±0.5          & 54.2±0.1          & 10.4±0.1                     & 65.8±0.1                     & 50.5±0.1                      & 43.6±0.1                     \\
PL         & 70.6±0.2          & 26.4±0.1          & 56.7±0.2          & 15.8±0.3                     & 72.6±0.2                     & 55.9±0.3                      & 49.6±0.2                     \\
SHOT       & 62.6±0.2          & 22.4±0.5          & 55.2±0.2          & 12.8±0.2                     & 62.8±0.4                     & 52.5±0.2                      & 44.7±0.3                     \\
FADA       & 60.7±0.1          & 22.0±0.3          & 53.2±0.2          & 9.4±0.1                      & 62.5±0.3                     & 50.0±0.2                      & 43.0±0.2                     \\
CPGA       & 66.7±0.2          & 27.8±0.1          & 54.3±0.1          & 12.7±0.2                     & 70.0±0.3                     & 51.9±0.2                      & 47.2±0.1                     \\
TransDA    & 64.9±0.1          & 22.7±0.2          & 54.1±0.2          & 11.8±0.2                     & 60.8±0.2                     & 49.6±0.1                      & 44.0±0.1                     \\
DECISION   & 60.5±0.2          & 20.6±0.5          & 56.2±0.1          & 12.6±0.2                     & 61.5±0.2                     & 53.6±0.4                      & 44.2±0.2                     \\
FADE       & 68.9±0.2          & 27.4±0.3          & 57.8±0.3          & 12.3±0.4                     & 72.4±0.2                     & 54.4±0.3                      & 48.9±0.3                     \\
KD3A       & 71.3±0.2          & 28.5±0.2          & 60.7±0.2          &15.8±0.2 & \multicolumn{1}{l}{73.4±0.1} & \multicolumn{1}{l|}{58.7±0.1} & \multicolumn{1}{l}{51.7±0.1} \\
Ours       & \textbf{74.0±0.2} & \textbf{30.0±0.1} & \textbf{62.4±0.1} & \textbf{19.0±0.1}   & \textbf{75.3±0.5}            & \textbf{62.0±0.2}             & \textbf{53.8±0.2}            \\ \hline
\end{tabular}
\end{table*}

\begin{table}[!thp]
\renewcommand\arraystretch{1.3}
\centering
\scriptsize      
\caption{Ablation study on OfficeHome. \emph{DA} and \emph{SM} indicate that the module of \emph{domain augmentation} and \emph{semantic matching} are disabled, respectively.}
\label{tb:ablationstudy}
\begin{tabular}{c|cccc|c}
\hline
w/o        & Art               & Clipart           & Product           & RealWorld         & Average           \\ \hline
SourceOnly & 77.6±0.4          & 60.4±0.5          & 84.8±0.4          & 85.4±0.4          & 77.1±0.4          \\ \hline
DA         & 79.9±0.2          & 64.5±0.3          & 88.5±0.1          & 87.0±0.3          & 80.0±0.2          \\
SM         & 79.0±0.3          & 64.1±0.6          & 87.9±0.1          & 86.9±0.2          & 79.5±0.2          \\
Ours       & \textbf{80.2±0.1} & \textbf{65.3±0.5} & \textbf{89.2±0.1} & \textbf{88.6±0.1} & \textbf{80.8±0.2} \\ \hline
\end{tabular}
\end{table}

{\bf OfficeHome.} OfficeHome is a benchmark dataset for domain adaptation and it consists of 15,500 images of 65 classes from four domains: Artistic (Ar), Clip Art (Cl), Product (Pr), and Real-world (Rw) images. This is a benchmark dataset for domain adaptation, with an average of around 70 images per class and a maximum of 99 images in a class. The images can be found typically in Home and Office settings.

{\bf OfficeCaltech.} Caltech-10 consists of pictures of objects belonging to 10 classes, plus one background clutter class. Each image is labeled with a single object. Each class contains roughly 40 to 800 images, while most classes have about 50 images, totaling around 9000 images. The size of the images are not fixed, with typical edge lengths of 200-300 pixels. 

{\bf PACS.} PACS is another popular benchmark for MSDA, which is composed of four domains (Art, Cartoon, Photo and Sketch). Each domain includes samples from 7 different categories, including a total of 9, 991 samples.

{\bf Breast Cancer.} Breast Cancer dataset includes 201 samples of one category and 85 samples of another category. The samples are described by 9 attributes, some of which are nominal and some are linear. 

\subsection{Comparison Baselines}
We compare FDAC with eleven state-of-the-art or representative approaches in terms of prediction accuracy. \emph{ResNet50} represents that the backbone is ResNet50, which is a popular deep architecture in CNNs. \emph{ResNet50} works in the source only manner. The only difference between \emph{R50-Ours} and our proposed method FDAC is that the backbone of \emph{R50-Ours} ResNet50. In \emph{R50-Ours}, we select the last layer for domain augmentation. Thus, both \emph{ResNet50} and \emph{R50-Ours} are CNNs-based, while the backbone of the left comparative methods are ViT-based. \emph{Source Only} is frequently used as a baseline to examine the advantage of domain adaptation methods. \emph{PL} is a pseudo-labeling approach in the source only manner, where the target domain trains a model with pseudo labels from the output of source classifiers. We use \emph{PL} to further prove the strong performance of ViT in feature extraction. For the above four methods, we change them into the federated setting. 

\emph{SHOT} \cite{liang2020we} only needs a well-trained source model and it aims to generate target data representations that can be aligned with the source data representations. \emph{DECISION} \cite{ahmed2021unsupervised} can automatically combine the source models with suitable weights where the source data is not available during knowledge transfer. \emph{TransDA} \cite{yang2021transformer} is based on Transformer and the corresponding attention module is injected into the convolutional networks. \emph{FADA} \cite{peng2019federated} designs a dynamic attention mechanism to leverage feature disentanglement to promote knowledge transfer. \emph{KD3A} \cite{feng2021kd3a} performs decentralized domain adaptation based on knowledge distillation and pseudo labeling, while it is also robust to negative transfer and privacy leakage attacks. \emph{CPGA} \cite{qiu2021source} first generates prototypes and pseudo labels, and then aligns the pseudo-labeled target data to the corresponding source avatar prototypes. \emph{FADE} \cite{hong2021federated} attempts to study federated adversarial learning to achieve goals such as privacy-protecting and autonomy.

\subsection{Implementation Details}
We implemented FDAC and the baseline methods using PyTorch \cite{paszke2017automatic}. We use the ViT-small with 16 $\times$ 16 patch size, pre-trained on ImageNet, as the ViT backbone. In each epoch, FedAvg \cite{mcmahan2017communication} is used to aggregate models after $r$ times of training. In our experiments, $r=1$. For model optimization, we set the Stochastic Gradient Descent (SGD) with a momentum of 0.9. The initial learning rate $\eta=10^{-3}$, which decays inversely with epochs. The batch size is 64, 128, or 256, depending on the actual number of samples in the current domain. About the parameters in Eq. (\ref{eq:obj}), we set $\lambda_1 = 1, \lambda_2=1$ for OfficeHome and OfficeCaltech, and $\lambda_1 = 0.2, \lambda_2=0.5$ for DomainNet, respectively.

\begin{figure*}[!t]
	\centering
	\includegraphics[height=5cm, width=16cm]{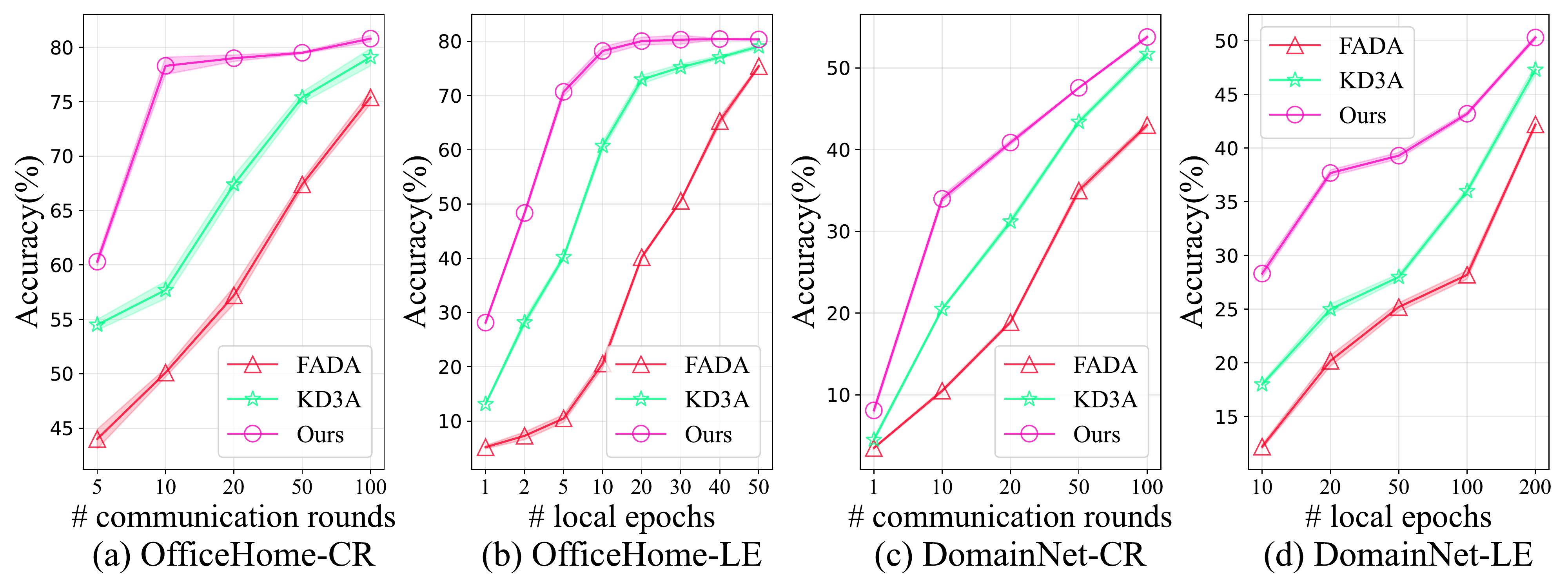}
	\caption{Comparison results under different numbers of communication rounds (CR) and local epochs (LE). }
	\label{fig:CommunicationEfficiency_LocalEpochs}
	\vspace{-0.5cm}
     \hspace{0.5cm}
\end{figure*}

\begin{figure}[!t]
	\centering
	\includegraphics[height=7cm,width=8cm]{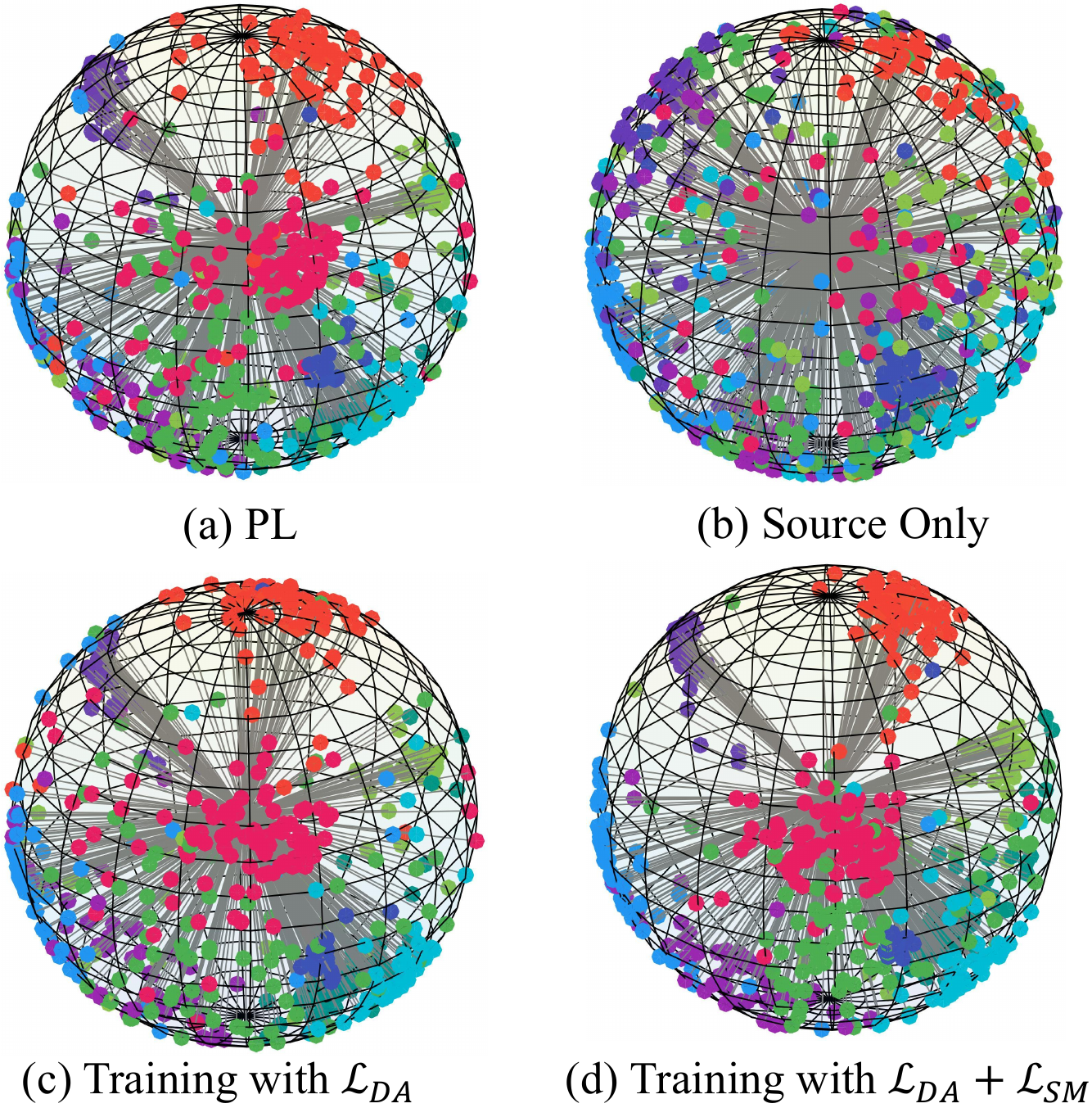}
	\caption{Feature visualization. (a) PL represents the policy of pseudo labeling. (b) indicates the result based on the source only manner. (c) represents that only domain augmentation is used in FDAC. (D) represents FDAC. Best viewed in color.}
	\label{fig:feas_viusal}
	\vspace{-0.5cm}
\end{figure}

\subsection{Experimental Results}
In FDA, there are multiple source domains and only one target domain. Thus, for each dataset, at one training time, only one sub-dataset is selected as the target domain while all the other sub-datasets are considered as the source domains. Take Table \ref{tb:OfficeHome} for example, the column \emph{Art} represents that Art is the target domain while Clipart, Product, and RealWorld are the source domains.

Table \ref{tb:OfficeHome} summarizes the results on OfficeHome. It is clear that FDAC outperforms the other methods in all sub-datasets. Notably, FDAC is much more effective on several sub-datasets such as Clipart and Product. The performance of FDAC is much better than \emph{R50-Ours}, representing that ViT plays an important role in feature extraction. 

The performance results on OfficeCaltech are summarized in Table \ref{tb:OfficeCaltech}. We note that FDAC also performs the best in most conditions. Compared to the results in Table \ref{tb:OfficeHome}, the performances of all the methods are relatively higher and the reason might be that this dataset is simpler than OfficeHome.

Experimental results on the dataset DomainNet are presented in Table \ref{tb:DomainNet}. This dataset is extremely challenging for two reasons. Firstly, the domain discrepancy in each adaption direction is important. Secondly, Too many categories (i.e., 345) make learning discriminative features much more challenging. \emph{R101-Ours} represents that the backbone is ResNet101, since this dataset is more complex than OfficeHome and OfficeCaltech. The performance of \emph{R101-Ours} is not bad and the reason might be that a complex CNNs can also be trained to extract adaptable features. FDAC outperforms all the comparative methods, indicating that domain-level augmentation and semantic matching can better enable domain adaptation in the federated setting.

\subsection{Further Analysis}

\subsubsection{Communication Efficiency}
Communication efficiency is an important indicator in the federated setting. To evaluate the communication efficiency, we train FDAC with different communication rounds $r$ and report the average accuracy on dataset OfficeHome and DomainNet. KD3A and FADA are selected as comparative methods. We set $r=$1, 2, 5, 10, and 20, representing that we synchronize models after $r$ rounds of training. Fig.\ref{fig:CommunicationEfficiency_LocalEpochs}.a-b shows the accuracy in each round during training. It is clear that the accuracy of all methods increases with the number of rounds, representing that FADA needs larger communication rounds for better performance. KD3A performs better than FADA, but it is still not so good as our method. For example, FDAC outperforms KD3A with more than 5\% accuracy, especially in the lower communication rounds (i.e., $r=5$). FDAC needs about half the number of communication rounds compared with KD3A. Moreover, FDAC is also robust to communication rounds and its accuracy only drops about 2\% when $r$ decreases from 100 to 10. In summary, our method is much more communication-efficient than the other methods. 

We also analyze the convergence property in FDAC and the results are displayed in Fig. \ref{fig:CommunicationEfficiency_LocalEpochs}.c-d. When the number of local training epochs is small, all methods perform poorly due to less training data. FDAC leads to the best convergence rate among the comparative methods. Moreover, we find that the other methods can hardly improve the performance of FDA with the ViT backbone. 

\subsubsection{Feature visualization}
To further investigate the feature distributions under our FDAC method, we randomly sample pixels on ViT-small based embedding from 10 categories on task $Clipart, Product, RealWorld \rightarrow Art$. We present the visualization under DA and SM, which are discussed in Eq. (\ref{eq:layer}) and Eq. (\ref{eq:semantic}), respectively. From Fig. \ref{fig:feas_viusal} we can get the following conclusions: (1) the policies of pseudo labeling and source only are not as good as the domain-augmentation module in FDAC; (2) the module of semantic matching can further improve knowledge transfer; (3) Both feature transferability and discriminability can be guaranteed in FDAC.

\subsubsection{Ablation study on Domain Augmentation and Semantic Matching}

To further analyze our approach FDAC, we conduct ablation experiments to fully investigate the effectiveness of different items as well as the sensitivity of hyper-parameters in the objective function. The loss elements in Eq. (\ref{eq:obj}) are jointly minimized to train the classifier. We disable one loss at each time and then record the result to evaluate its importance on OfficeHome. The results are displayed in Table \ref{tb:ablationstudy}. For all the sub-datasets, it is clear that each loss item is necessary to guarantee performance, indicating that both domain augmentation and semantic matching are important in FDAC. 

Take OfficeHome for example, $\lambda_1$ and $\lambda_2$ are similar in sensitivity.

\begin{table}[!t]
\centering
\scriptsize
\caption{Parameter sensitivity of $\lambda_1 \text{and} \lambda_2$ on OfficeHome}
% \caption{Parameter sensitivity of $\lambda_1 \text{and} \lambda_2$.}
\label{tb:sensitity}
\begin{tabular}{c|cccc}
\hline
            & Art  & Clipart & Product & RealWorld \\ \hline
$\lambda_1=0.1$ & 79.2 & 64.4    & 88.0    & 87.7      \\
$\lambda_1=0.5$ & 79.0 & 64.5    & 88.5    & 88.1      \\
$\lambda_1=1.0$ & 80.2 & 65.3    & 89.2    & 88.6      \\
$\lambda_1=1.5$ & 78.9 & 63.9    & 88.4    & 87.9      \\ \hline
$\lambda_2=0.1$ & 79.0 & 63.4    & 87.7    & 87.8      \\
$\lambda_2=0.5$ & 79.3 & 64.5    & 88.3    & 88.1      \\
$\lambda_2=1.5$ & 79.6 & 64.6    & 88.5    & 88.3      \\ \hline
\end{tabular}
\vspace{-0.3cm}
\end{table}

\subsubsection{Domain Augmentation based on Latent Manipulation}
\begin{figure}[!t]
	\centering
	\includegraphics[width=8cm]{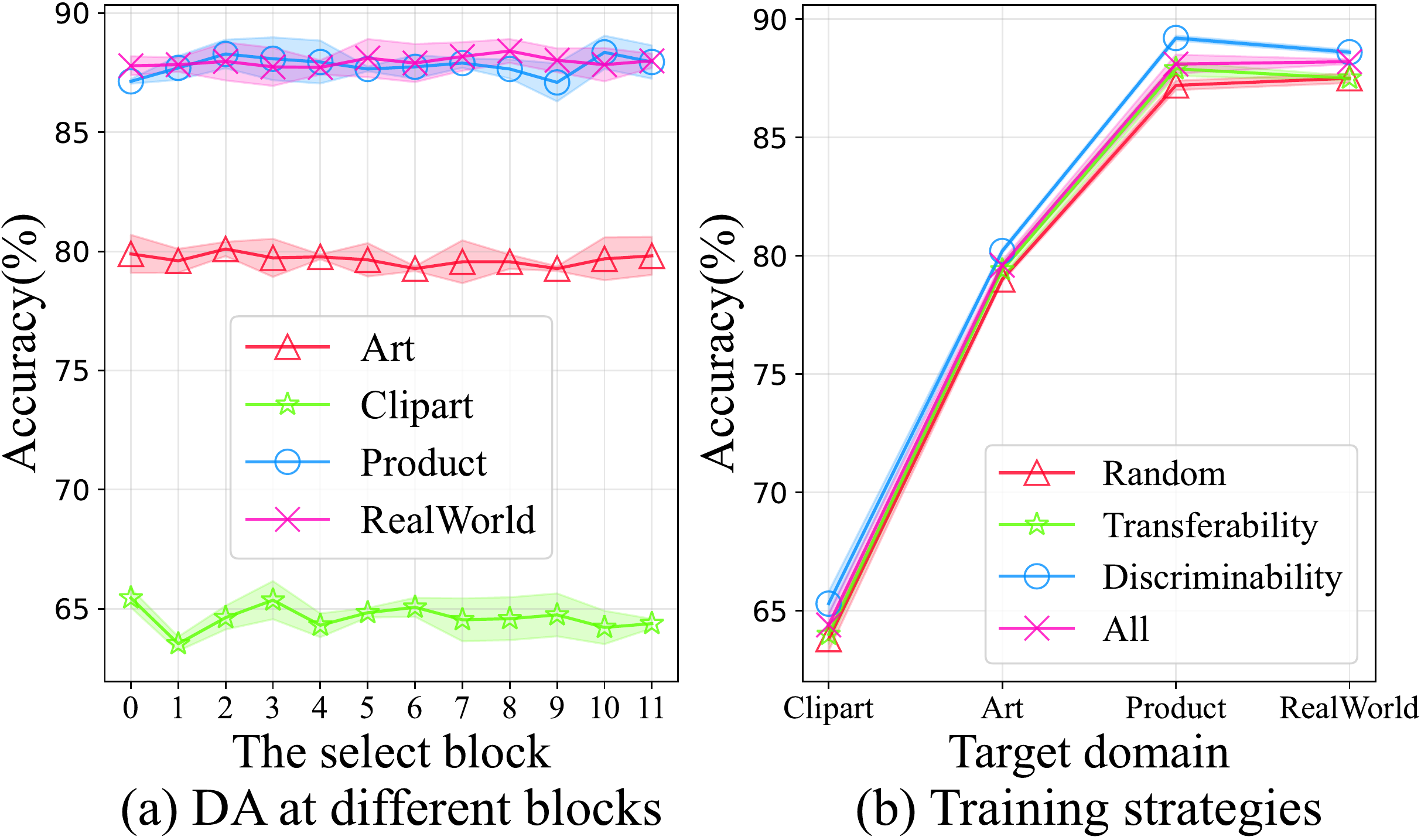}
	\caption{Analysis of the hidden manipulation (Domain augmentation, DA) of ViT architecture. }
	\label{fig:ViT_Architecture}
	\vspace{-0.5cm}
\end{figure}

Table \ref{tb:ablationstudy} indicates that the policy of domain augmentation can enhance domain adaptation in the federated setting, thus, it is interesting to investigate which block in ViT is the most important to the performance of FDAC. We choose one block at one time to examine and the result of upon OfficeHome is displayed in Fig. \ref{fig:ViT_Architecture}.a. It can be observed that the best block for domain augmentation varies from one task to another.

In order to further exploit the importance of different blocks in domain augmentation, we use another four strategies to select the block: \textit{Transferability} means to select the previous blocks of ViT; \textit{Discriminability} means to select the later layers of ViT; \textit{Random} represents that the block is randomly selected; \textit{All} represents that all blocks are selected. The result in Fig.\ref{fig:ViT_Architecture}.b demonstrates that it is better to select \textit{Discriminability} blocks for domain augmentation. The reason might be that aligning the later blocks is better to keep the transferability of features since those blocks are relatively more discriminative.

\subsubsection{The Advantage of Domain Augmentation}
The policy of domain augmentation in FDA is to extract transferable features, thus we investigate the advantage of this policy with two other representative techniques, i.e., Mixup \cite{zhang2017mixup} and SSRT \cite{sun2022safe}. Mixup combines two samples linearly. Formally, let $x_i$ and $x_j$ be two target samples, and $y = \mathcal{G}(x)$ be the model classifier predictions. We mix target samples with a designed weight $\lambda$ sampled from a Beta distribution by a parameter $\beta$. The data is mixed at domain-level and the augmented data $\left( \widetilde{x}, \widetilde{y}\right)$ can be computed by:

\begin{equation}
\label{eq:mixup1}
\left\{
             \begin{array}{lr}
             \lambda  \sim Beta(\beta, \beta), &\\
             \widetilde{x} = \lambda x_i  + (1 - \lambda) x_j, & \\
             \widetilde{y} = \lambda y_i + (1 - \lambda) y_j. &   
             \end{array}
\right.
\end{equation}
% \end{center}
The corresponding optimal function is defined as:
\begin{equation}
    \label{eq:mixup2}
    \begin{aligned}
    \mathcal{L}_m = - \underset{\widetilde{x} \sim D_T}{\mathbb{E}} \widetilde{y} \log {\mathcal{G}(\widetilde{x})}.
    \end{aligned}
\end{equation}

\begin{figure}[!t]
	\centering
	\includegraphics[width=8cm]{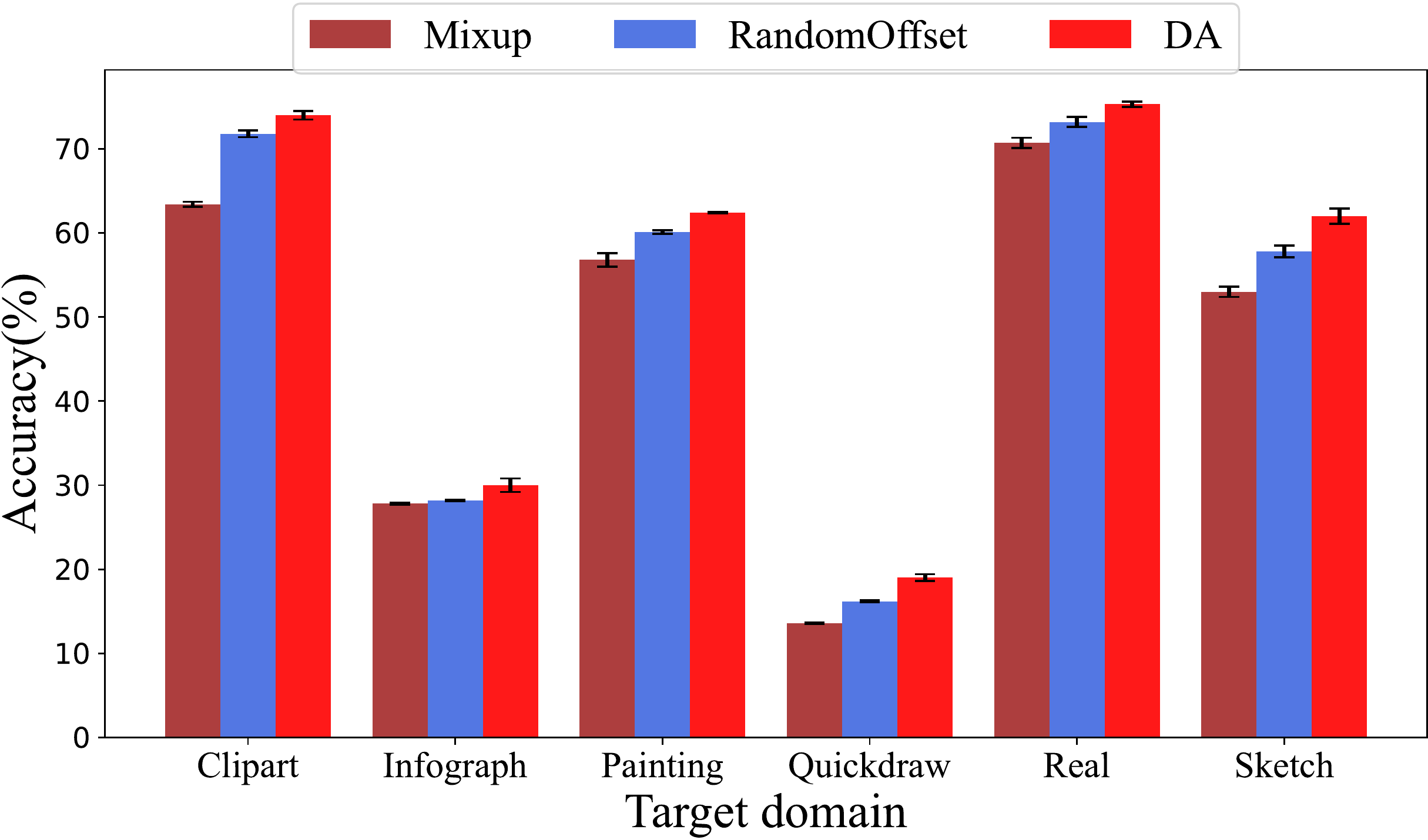}
	\caption{The advantage of domain augmentation (\emph{DA}) in FDAC. }
	\label{fig:Augmentation_methods}
	\vspace{-0.5cm}
\end{figure}

SSRT \cite{sun2022safe} first adds random offsets to the latent token sequences of target sample, and then minimizes the discrepancy of the model's prediction between the original and augmented data by Kullback Leibler (KL) divergence \cite{sugiyama2007direct}. Let $b_x^l$ be the latent representation of original input $x$ and $b^l_{xr}$ be the augmented representation which adds an offset. The augmented data $\widetilde{b}^l_{x}$ can be obtained by:
\begin{equation}
    \label{eq:randomoffset1}
    \begin{aligned}
    \widetilde{b}^l_{x} = b_x^l + \alpha \left[b_x^l - b^l_{xr} \right]_{\times},
    \end{aligned}
\end{equation}
where $\alpha$ is a scalar parameter and $[\cdot]_{\times}$ means no gradient backpropagation. Let $p_x$ and $\widetilde{p}_x$ be the model predictions corresponding to $b_x^l$ and $\widetilde{b}^l_{x}$, respectively. Then, the loss function can be defined as:
\begin{equation}
    \label{eq:randomoffset2}
    \begin{aligned}
    \mathcal{L}_r = \underset{\widetilde{x} \sim D_T}{\mathbb{E}} p_x \log \left(\frac{p_x}{\widetilde{p}_x} \right).
    \end{aligned}
\end{equation}

We use $\mathcal{L}_m$ and $\mathcal{L}_r$ to replace $\mathcal{L}_{DA}$ in Eq. (\ref{eq:obj}). For all tasks, $\alpha$ and $\beta$ are set to be 1 and 0.2, respectively. Fig. \ref{fig:Augmentation_methods} presents the results on the dataset DomainNet based on the ViT-base backbone. It is clear that the domain augmentation policy in FDAC is better than the two other data augmented policies, and the reason might be that the complementarity from source domains to the target domain is considered in FDAC.

\section{Conclusions}
In this paper, we propose a novel approach, namely FDAC, to address federated domain adaptation via contrastively transfer knowledge from different source models to the target model. Firstly, we manipulate the latent architecture of ViT to further extract transferable features among domains, where the data is contrastively augmented at domain-level thus the data diversity of the target domain is also enhanced. Secondly, we generate prototypes for each source domain and high-quality pseudo labels for the target domain to bridge the domain discrepancy based on contrastive learning. In this way, both feature transferability and discriminability can be guaranteed and the knowledge can be leveraged to adapt across models. 

Extensive experiments on different real classification and segmentation tasks demonstrate the outstanding performance of FDAC in federated domain adaptation, and the communication efficiency is simultaneously guaranteed. Furthermore, the comparative results also indicate that our domain augmentation under ViT is better than existing ViT-based augmentation methods.

{\small
\bibliographystyle{IEEEtran}
\bibliography{egbib}
}

% \begin{thebibliography}{1}
% \bibliographystyle{IEEEtran}

\end{document}